\documentclass[sigconf]{acmart}

\AtBeginDocument{
  }

\setcopyright{none}

\author{Yuebo Luo$^{1}$, Shiyang Li$^{1}$, Junran Tao$^{2}$, Kiran Thorat$^{3}$, Xi Xie$^{3}$, Hongwu Peng$^{3}$, Nuo Xu$^{1}$, Caiwen Ding$^{1}$, Shaoyi Huang$^{2}$}

\affiliation{\fontsize{11pt}{11pt}\selectfont \institution{$^1$University of Minnesota, Twin Cities, 
$^2$ Stevens Institute of Technology,
$^3$ University of Connecticut.}
 \country{USA}}

\affiliation{\fontsize{9pt}{9pt}\selectfont
$^{1}$\{luo00466, li004074, xu001536, dingc\}@umn.edu \\$^{2}$\{jtao11, shuang59\}@stevens.edu\\
$^{3}$\{kiran\_gautam.thorat, xi.xie, hongwu.peng\}@uconn.edu 
\country{}
}

\usepackage{multirow}
\usepackage{algorithm}
\usepackage{amsmath}
\usepackage{algpseudocode}
\usepackage{graphicx}
\usepackage{subcaption} 
\usepackage{lipsum} 
\usepackage{arydshln}
\usepackage{pifont}

\begin{document}

\title{DR-CircuitGNN: Training Acceleration of Heterogeneous Circuit Graph Neural Network on GPUs}
\date{}

\begin{abstract}
The increasing scale and complexity of integrated circuit design have led to increased challenges in Electronic Design Automation (EDA). Graph Neural Networks (GNNs), have emerged as a promising approach to assist EDA design as circuits can be naturally represented as graph.
While GNNs offer a foundation for circuit analysis, they often fail to capture the full complexity of EDA designs.
Heterogeneous Graph Neural Networks (HGNNs) can better interpret EDA circuit graphs as they capture both topological relationships and geometric features.
However, the improved representation capability comes at the cost of even higher computational complexity and processing cost due to their serial module-wise message-passing scheme, creating a significant performance bottleneck.
In this paper, 
we propose~\textbf{DR-CircuitGNN}, a fast GPU kernel design by 
leveraging row-wise sparsity-aware Dynamic-ReLU 
and optimizing SpMM kernels during heterogeneous message-passing to accelerate HGNNs training on EDA-related circuit graph datasets. To further enhance performance, we propose a parallel optimization strategy that maximizes CPU-GPU concurrency by concurrently processing independent subgraphs using multi-threaded CPU initialization and GPU kernel execution via multiple cudaStreams.
Our experiments show that on three representative CircuitNet designs (small, medium, large), the proposed method can achieve up to 3.51 $\times$ and 4.09 $\times$ speedup compared to the SOTA for forward and backward propagation, respectively.
On full-size CircuitNet and sampled Mini-CircuitNet, our parallel design enables up to $2.71\times$ speed up over the official DGL implementation cuSPARSE with negligible impact on correlation scores and error rates.
\end{abstract}


\maketitle

\pagestyle{plain}

\section{Introduction}

The growing demand for semiconductor Integrated Circuits (ICs) and the slowdown of Moore’s Law have driven increased interest in leveraging Machine Learning (ML) to enhance Electronic Design Automation (EDA) tools and processes~\cite{lopera2021survey,alrahis2021gnn,bustany2015ispd}. 
Given that electronic circuits inherently consist of interconnected components, they are well-suited for graph-based representations. Consequently, Graph Neural Networks (GNNs) have been effectively employed for tasks such as predicting links between circuit components and optimizing circuit placement, demonstrating substantial improvements in critical design phases\cite{mirhoseini2021graph,cheng2021joint,kirby2019congestionnet,ghose2021generalizable,alawieh2020high}. Among various GNNs, Heterogeneous GNNs (HGNNs) are particularly significant for circuit graph modeling, as they capture both topological relationships (e.g., wire routing) and geometric features (e.g., cell placements), offering a unified framework for circuit design optimization. 

\begin{figure}[t]  % Changed from figure to figure*  
        \centering
        \includegraphics[width=0.95\linewidth]{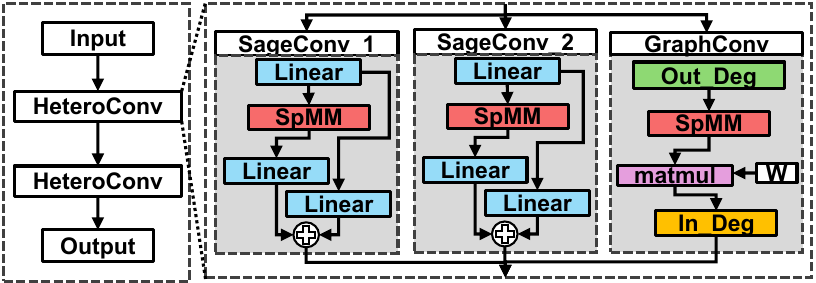}
        \caption{HGNN architecture.  A HeteroConv block comprises three GNN modules (i.e., two SageConv modules and one GraphConv module).}
        \label{fig:Heteroconv}
        \vspace{-10pt}
    % \hfill
\end{figure}

The increasing complexity of modern IC design has caused EDA datasets \cite{sigda2012,bustany2015ispd} to expand exponentially, with training sets now including tens of thousands of designs and heterogeneous graphs reaching terabyte-scale sizes\cite{10158384,2024circuitnet}. This dual expansion in scale and complexity introduces efficiency bottlenecks in HGNN training.

\begin{figure}[ht]
    \centering
        \includegraphics[width=0.5\textwidth]{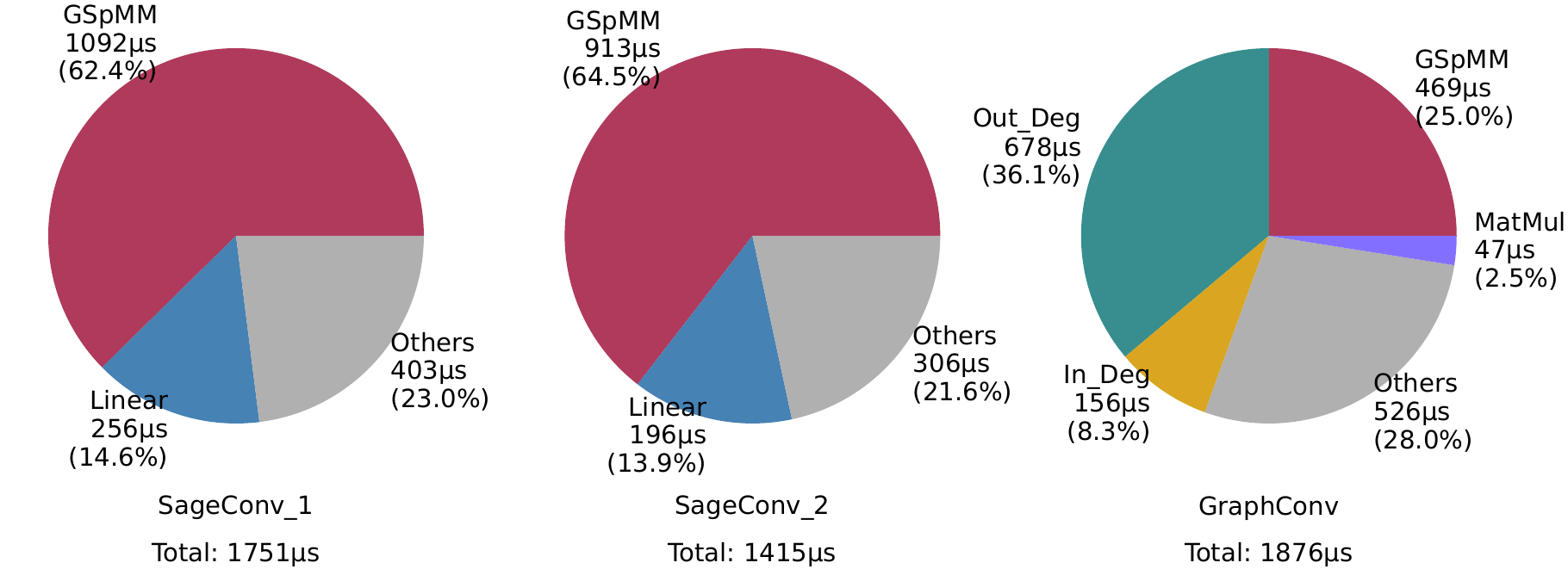}
        \caption{Three modules in one HeteroConv layer's training time breakdown on the CircuitNet dataset \cite{10158384}.}
        \label{fig:HeteroTimeDist}
        \vspace{-0.5cm}
\end{figure}

Through comprehensive profiling analysis, we have identified three major bottlenecks. (i) At the kernel level, Sparse Matrix-Matrix Multiplication (SpMM) serves as the core message-passing mechanism in HGNNs, as illustrated in the architecture shown in Figure~\ref{fig:Heteroconv}. Crucially, SpMM is recognized as the primary performance bottleneck~\cite{gilmer2017neural}. SpMM is a fundamental linear algebra operation where a sparse matrix (matrix with mostly zero elements) is multiplied by a dense matrix (matrix with mostly non-zero elements), producing a dense output matrix. The architecture in Figure~\ref{fig:Heteroconv} consists of three sub-modules (from left to right: SageConv, SageConv, and GraphConv), each handling a specific type of edge, respectively. Figure~\ref{fig:HeteroTimeDist} underscores the performance impact, revealing that during inference, SpMM accounts for a significant portion of the forward overhead: approximately 62.4\%, 64.5\%, and 25.4\% of the runtime within these respective modules. Furthermore, the backward-pass SpMM also contributes significantly to the overall overhead, highlighting the pervasive impact of SpMM inefficiencies across both forward and backward propagation.

(ii) At the workload schedule level, the training pipeline suffers from inefficiencies due to workload imbalance and underutilized parallelism. The irregular distribution of node degrees -- where certain nodes exhibit significantly higher connectivity than others -- causes a critical performance bottleneck~\cite{geng2020awb}. 

(iii) Furthermore, the system design and optimization of heterogeneous graphs remain underexplored. For example, current GNN libraries and implementations, e.g., Deep Graph Library (DGL)~\cite{dgl}, sequentially process different subgraphs 
despite their computational independence until the final aggregation phase. This coarse-grained scheduling incurs unnecessary synchronization overhead, leading to a significant underutilization of GPU resources. These inefficiencies collectively hinder the scalability of HGNNs for large-scale EDA-related applications.

These challenges underscore the need for efficient kernel optimization, an optimized workload mapping method, and better scheduling strategy for training an HGNN using large-scale heterogeneous circuits. 
To address these challenges, in this paper, 
we propose DR-CircuitGNN, an acceleration framework that leverages row-wise sparsity-aware \underline{D}ynamic-\underline{R}eLU for heterogeneous \underline{circuit} \underline{GNN} training. 
\\

The contributions of the paper are
summarized as follows:
\begin{itemize}
\item We introduce a novel inter and intra subgraph-level Dynamic ReLU (D-ReLU) mechanism specifically designed for heterogeneous graphs. By dynamically thresholding node embeddings on a per-row basis, this method injects balanced row sparsity during training, which effectively reduces workload imbalance on GPU.
\item We design a specialized forward-pass SpMM kernel (DR-SpMM forward) that targets the irregular adjacency patterns in circuit sub-graphs. By mapping row-sparsity directly onto each edge type and node type, DR-SpMM significantly accelerates the message-passing stage under heterogeneous circuit constraints.
\item For the backward pass, we introduce a DR-SpMM backward kernel customized to circuit graph gradients. This kernel promptly leverages column-major neighbor indexing -- explicitly accommodating net-to-cell and cell-to-net connections -- to achieve fast and efficient gradient aggregation in large-scale circuit training.
\item We develop a parallel scheduling mechanism that concurrently processes all subgraph updates originating from the same circuit design. We parallelize the processing of subgraphs both on the CPU side and GPU side, with multi-threads and multiple \textit{cudaStream} separately. Such maximizes CPU-GPU concurrency and GPU resource utilization, further boosting overall training efficiency.
\end{itemize}

Experimental results show that DR-CircuitGNN achieves 3.21$\times$ and 2.75$\times$ speedup on forward propagation, as well as 3.51$\times$ and 4.09$\times$ speedup on backward propagation with no accuracy loss in congestion prediction tasks when compared to DGL~\cite{dgl} and GNNAdvisor ~\cite{wang2021gnnadvisor}, respectively. Furthermore, our optimal parallel message-passing pipeline yields on average $2.69\times$ forward and $2.70\times$ backward acceleration over cuSPARSE~\cite{cusparse}, as well as $11.06\times$ forward and $11.07\times$ backward acceleration over GNNA.

\section{Background and Motivation}
\subsection{Message-passing in Graph Neural Networks}

The message-passing (MP) mechanism in GNNs operates by updating node embeddings based on the structural information encoded in the adjacency matrix. This process can be formulated as:

\begin{equation}
    \underbrace{\mathbf{X}^{(l+1)}}_{N \times F} = \text{UPDATE}\left(
        \text{MSG}\left(\underbrace{\mathbf{A}}_{N \times M}, \underbrace{\mathbf{X}^{(l)}}_{N \times F}\right), 
        \text{AGG}\left(\mathbf{X}^{(l)}\right)
    \right)
\end{equation}
where $l$ denotes the $l$-th number of iterations of the message-passing process, the message function $MSG$ defines the transformation of node features from source to destination nodes, while the aggregation function $AGG$ specifies how the transformed features are processed at destination nodes. In homogeneous graphs, where connections exist only between nodes of the same type, $N=M$ represents the number of nodes serving as both sources and targets. However, heterogeneous graphs typically involve $M$ source nodes and $N$ target nodes, where $N \neq M$ due to varying node type quantities. Owing to the highly sparse nature of the adjacency matrix and the dense matrix of node embedding, the product between $A$ and $X$ becomes the Sparse Matrix Multiplication (SpMM), which is the backbone of the message-passing process.

\subsection{Representing Circuit Design with Heterogeneous Graph Neural Networks}

The circuit graph construction process is depicted in Figure \ref{fig:topo_geom}. CircuitNet first directly extracts physical encoded features from the layout in $(a)$ and then builds topological links between cells and nets, which can be read from the netlist using the estimation method \cite{xie2022preplacement}, as shown in step $(b)$. Meanwhile, in step $(c)$, a shifting window technique is applied to capture the geometrical connectivity among cells \cite{liu2021swin}. 
Finally, in step $(d)$, the circuit graph is designed with topological and geometrical links. 

As the core of the CircuitNet\cite{10158384}, circuit designs can be effectively represented through a combination of heterogeneous graph structures, incorporating both topological and geometric information. This representation of the dataset exhibits several distinctive characteristics:

\begin{figure}[t]
    \centering   \includegraphics[width=1.0\linewidth]{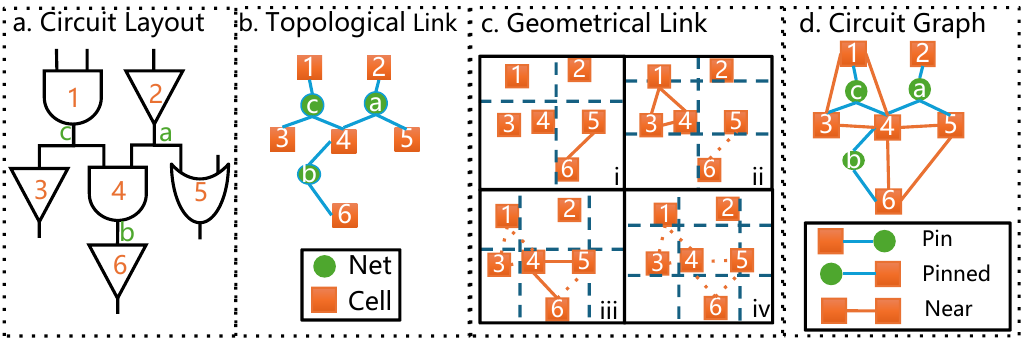}
    \caption{Circuit graph generation process: (a) Circuit Layout; (b)  Topological Link; (c) Geometrical Link; (d) Circuit Graph.
    % \textcolor{red}{explain heterogeneous graph here}
    }
    \label{fig:topo_geom}
    \vspace{-0.45cm}
\end{figure}

1) The CircuitNet dataset organizes raw data by design units, with each design partitioned evenly in general to keep roughly 10,000 nodes per graph: $Partition(\mathcal{D}) = \{G_1, G_2, \ldots, G_n\}$;

2) Under each design, each graph $G_i$ incorporates three edge types ($pin$, $pinned$, and $near$) representing interactions between two node types ($cell$ and $net$): $For$ $graph:\space   G_i = (V_i, E_i)$, there are $V_i = V_i^\text{cell} \cup V_i^\text{net}$ and $V_i^\text{cell} \cap V_i^\text{net} = \emptyset$, with $E_i = E_i^\text{near} \cup E_i^\text{pin} \cup E_i^\text{pinned}$;

3) $pins$ and $pinned$ are edges starting from $cell$ to 'net' and the opposite way, respectively, making their adjacency matrices the transposition of each other; $near$ describes the geometrical links between cells. With these various node types and edge types within an individual graph, the graphs become 'heterogeneous', which is: $E_i^{near}     \subseteq V_i^{cell} \times V_i^{cell} $ as cell-to-cell edge, $E_i^{pin}     \subseteq V_i^{cell}\times V_i^{net} $ as cell-to-net edge, and $ E_i^{pinned}   \subseteq V_i^{net} \times V_i^{cell} $ as net-to-cell edge.

The cardinality of node sets is defined as: $ |V_i^{cell}| = N_i $ and $ 
    |V_i^{net}| = M_i$, annotating the number of cells and nets, respectively.

\begin{figure}
    \centering
    \includegraphics[width=1.0\linewidth]{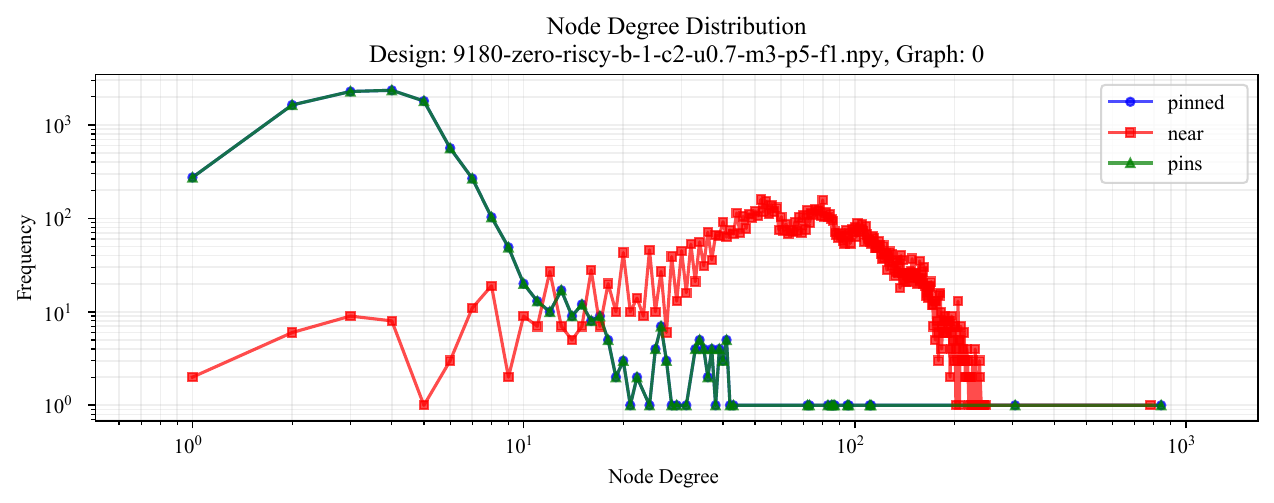}
    \caption{Node degree distribution of three subgraphs, each corresponds to the edge type of $pins$, $near$ and $pinned$ respectively. The subgraphs are from an example of CircuitNet\cite{10158384}.
    % \textcolor{red}{add more description to make this figure more clearly explained, and make it shorter vertically.}
    }
    \vspace{-0.13in}
    \label{fig:edge_near_pins}
\end{figure}

In the post-partitioning stage, each circuit graph contains 5,000 to 10,000 nodes of both types, resulting in varying neighbor densities across different edge types. As illustrated in Figure~\ref{fig:edge_near_pins}, the $near$ edges exhibit a concentrated distribution of neighbor counts per node, peaking around 50 neighbors and then rising to slightly over 250 per node. Whereas, $pins$ and $pinned$ edges show a more sparse connectivity pattern, with neighbor counts per node concentrated below 100. Combined with the observation of the varying embedding dimensions between $cell$ and $net$ nodes, this necessitates careful consideration with respect to SpMM workloads derived from adjacency matrices and node embeddings. This heterogeneous structure demands specialized
% activation pruning 
non-linear function with sparsification
and SpMM acceleration techniques to optimize the message-passing process effectively.

\subsection{Motivation}

\subsubsection{\textbf{SpMM is the bottleneck}}

% \textcolor{red}{add observation here @yuebo}
As the profiler result given in Figure \ref{fig:HeteroTimeDist} shows, SpMM methods take up the majority of the runtime during heterogeneous message-passing. While the commonly observed power-law distribution of numbers of neighbors per node \cite{geng2020awb} always brings out "evil rows", incurring serious workload imbalance when trying to process node-level neighbor aggregation in parallel \cite{wang2021gnnadvisor}, which greatly lowers the efficiency of SpMM methods. Following the observation, Figure \ref{fig:edge_near_pins} shows that a typical circuit graph contains multiple edge types, which leads various adjacency matrices, and all of them generally are subject to the power-law distribution with spiking at different regions, the edge type $near$ surges in neighbor numbers around 100 per node, while the other two edges sharing the same connection but different directions, $pins$ and $pinned$ concentrate at as low as 3 and 4, meaning not only the "evil rows" obviously exist only in one adjacency matrix, but also, varied groups of "evil rows" across all adjacency matrices need to be handled at one time, posing serious challenge to address the inherent bottleneck of SpMM.

\textbf{Workload Imbalance by Graph characteristics should be addressed.}
As we delved into implementing message-passing in terms of SpMM for bottleneck solutions. It is noticed that SpMM utilizes row-wise product \cite{wang2021gnnadvisor} to finish the neighbor aggregation. This method, on one hand, processes extremely sparse adjacency matrix serially by iterating on the indices of either Compressed Sparse Row (CSR) or the Compressed Sparse Column (CSC), while loading the relative neighbor node embedding indexed by the fetched neighbor connection (edge) using parallelized warps of threads. 

$$\mathcal{W}_i = |\mathcal{N}(i)| \times D$$
$\mathcal{W}_i$ represents the workload for node $i$, $|\mathcal{N}(i)|$ denotes the number of neighbors for node $i$, and $D$ is the dimension of the node embeddings.

For the maximum number of row-wise products computed in parallel, considering workload imbalance due to "evil rows":
$$\mathcal{P}_{max} = \min\left(\frac{T_{avail}}{\max_i |\mathcal{N}(i)| \times D}, V\right)$$
Where  $\mathcal{P}_{max}$ is the maximum number of row-wise products computable in parallel, $T_{avail}$ represents the total available parallel threads, $\max_i |\mathcal{N}(i)|$ captures the "evil row" \cite{geng2020awb} problem (the node with maximum neighbors), $V$ is the total number of nodes in the graph. 
This shows that workload is a crucial metric defining the bottleneck of efficiency of a row-wise product SpMM kernel. 
Combined with our observation, it is seriously needed that a dynamic, adaptive approach to allocate a suitable proportion of hardware resources in terms of threads, so that not only for in-edge message-passing, SpMM kernels suffer less from waiting for the evil rows to finish their row-wise product to synchronize, cutting down the tail lag and achieving overall shorter runtime.

\textbf{Irregular Memory Access Exacerbated in HGNN.}
In order to reduce the workload imbalance, one straightforward measure is to promptly reduce the number of non-zero node embeddings per neighbor $D$ that should be passed to the current node. 

This can be implemented through 
non-linear function applied to fully dense embeddings before the SpMM-based neighbor aggregation, such as Rectified Linear Unit (ReLU) \cite{agarap2019deeplearningusingrectified}, mapping all negative neurons to zeros in the output. However, owing to the random, non-deterministic distribution of values of embedding, even its threshold-aware variant FATReLU \cite{10.5555/3524938.3525451} easily leaves irregular, unbalanced sparsity. This irregular sparsity makes the CUDA kernel perform an irregular memory access pattern, harming the parallelism of the GPU. 
SpMM kernels do not have the pre-knowledge of non-zero distribution in the sparsified embeddings, nor do they have a fully dynamic scheduling scheme that perfectly processes randomly and irregularly distributed non-zeros without extra cost. Such a scenario calls for an adaptive, dynamic, non-linear function producing regular sparsity to enable a more efficient SpMM.

\begin{figure*}[h]
    \centering
    \includegraphics[width=1.0\linewidth]{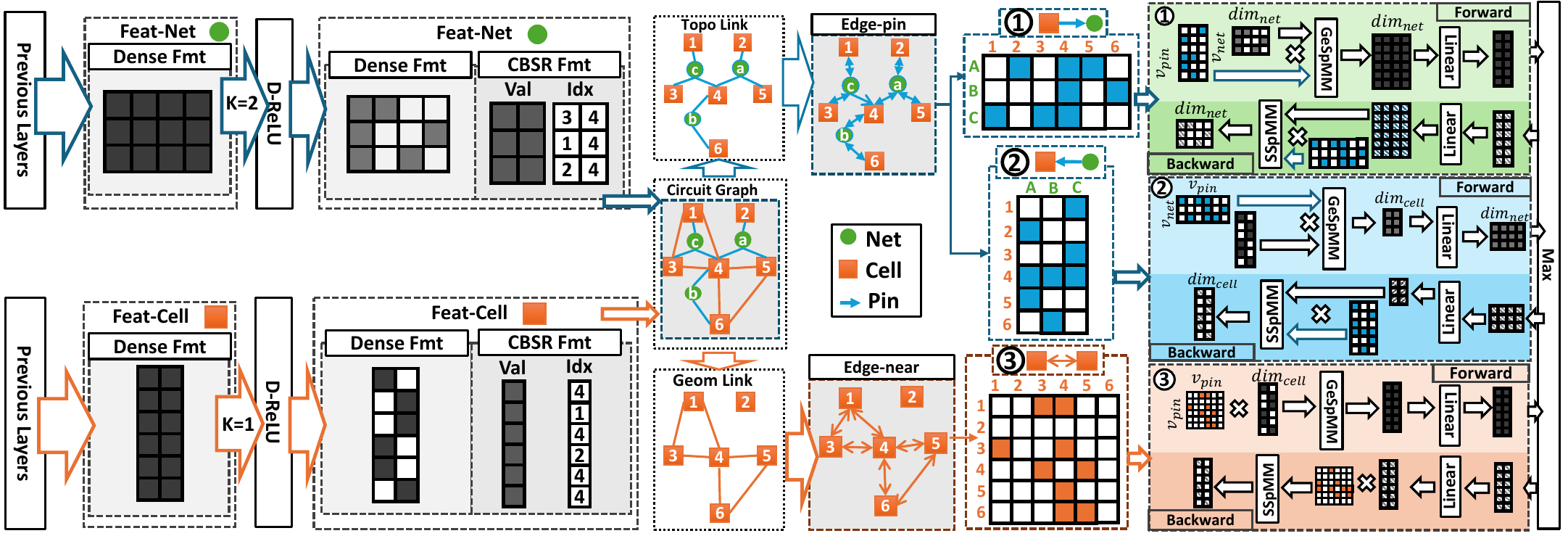}
    \caption{The overview of Message Passing of one layer of heterogeneous Graph Neural Network on circuit graph dataset.}
    \label{fig:Hetero-SpMM}
    % \vspace{-10pt}
\end{figure*}
\subsubsection{\textbf{HGNNs Need More parallelism}}

The sequential message-passing scheme currently adpoted by DGL and PyG keeps the whole-graph dataflow to wait for all types of subgraphs to be updated. However, the computing of updating of different types of subgraphs is totally independent, and each single workload most of times cannot full fill GPU computing resources. Furthermore, the frequent synchronization after each updating can significantly harm the GPU parallelism.
Therefore, such a sequential manner has caused a lot of waste of GPU time and resources. While some recent approaches work on building a parallel HGNN framework, their focus is confined to the graph reconstruction of semantic graphs, with designs of accelerators dedicated to their reconstruction methods \cite{10.1145/3649329.3656540,10510500,10650951}, where system optimization from the scheduling aspect is absent. Considering the extremely large scale of the EDA dataset, parallel processing for subgraphs can further improve the performance of HGNN.

\section{Design of DR-CircuitGNN}

In order to address our motivations, we design a \textit{Dynamic ReLU}(D-ReLU) non-linear function with sparsification for GNN 
, a framework of SpMM kernels based on output from D-ReLU, with heterogeneous forward and backward pass functionality, and a parallel pipeline enabling parallelized message-passing for all edges to accelerate the HGNN training flow fundamentally.
The overview of our proposed DR-CircuitGNN framework is shown in Figure \ref{fig:Hetero-SpMM}.
DR-CircuitGNN first accommodates varying embedding dimensions across node types, with different K-values applied to $net$ and $cell$ nodes' embedding to generate type-specific values and indices.
After the D-ReLU's sparsification by the K-values, in the heterogeneous message-passing phase, node embeddings are passed along the corresponding edges to perform forward \textit{Dynamic ReLU Sparse Matrix Multiplication} (DR-SpMM) using relative indices generated by D-ReLU. With a pipeline design, all three edges are capable of message-passing in parallel until the $cell$ node merges its updated embeddings transmitted from two types of edges, $pinned$ and $near$. When it comes to the backward pass, DR-CircuitGNN reuses the indices gain from the D-ReLU to apply Sampled-Sparse Matrix Multiplication (SSpMM) on the incoming gradient to compute and pass the gradients backward to the source nodes, finishing one whole training cycle of DR-CircuitGNN. 

\subsection{Hetero-Dynamic ReLU and DR-SpMM}

\begin{figure*}
    \centering
    \includegraphics[width=1.0\linewidth]{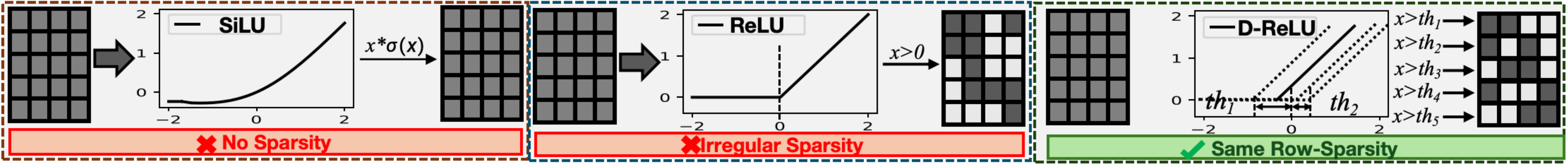}
    \caption{D-ReLU (ours) vs. other non-linear activations, including SiLU (leftmost) and ReLU (middle), } where D-ReLU generates row-wise balanced sparsity with top-k elements preserved in each row.
    \label{fig:D_ReLU}
    % \vspace{-10pt}
\end{figure*}

To facilitate efficient SpMM, we introduce a \textit{Dynamic ReLU} operator that selectively preserves the most significant elements of node embeddings through row-wise binary search. This approach differs fundamentally from conventional non-linear activations in GNNs, as illustrated in Figure~\ref{fig:D_ReLU}, from left to right, there are illustrations of Sigmoid Linear Unit (SiLU) \cite{elfwing2017sigmoidweightedlinearunitsneural}, ReLU, and our D-ReLU, respectively. While SiLU employs a sigmoid-based function ($x*\sigma(x)$) that maintains density, and standard ReLU \cite{agarap2019deeplearningusingrectified} introduces irregular sparsity through zero-thresholding, our D-ReLU implements dynamic, row-wise thresholding to achieve balanced sparsity patterns.
\begin{figure*}
    \centering
    \includegraphics[width=1.0\linewidth]{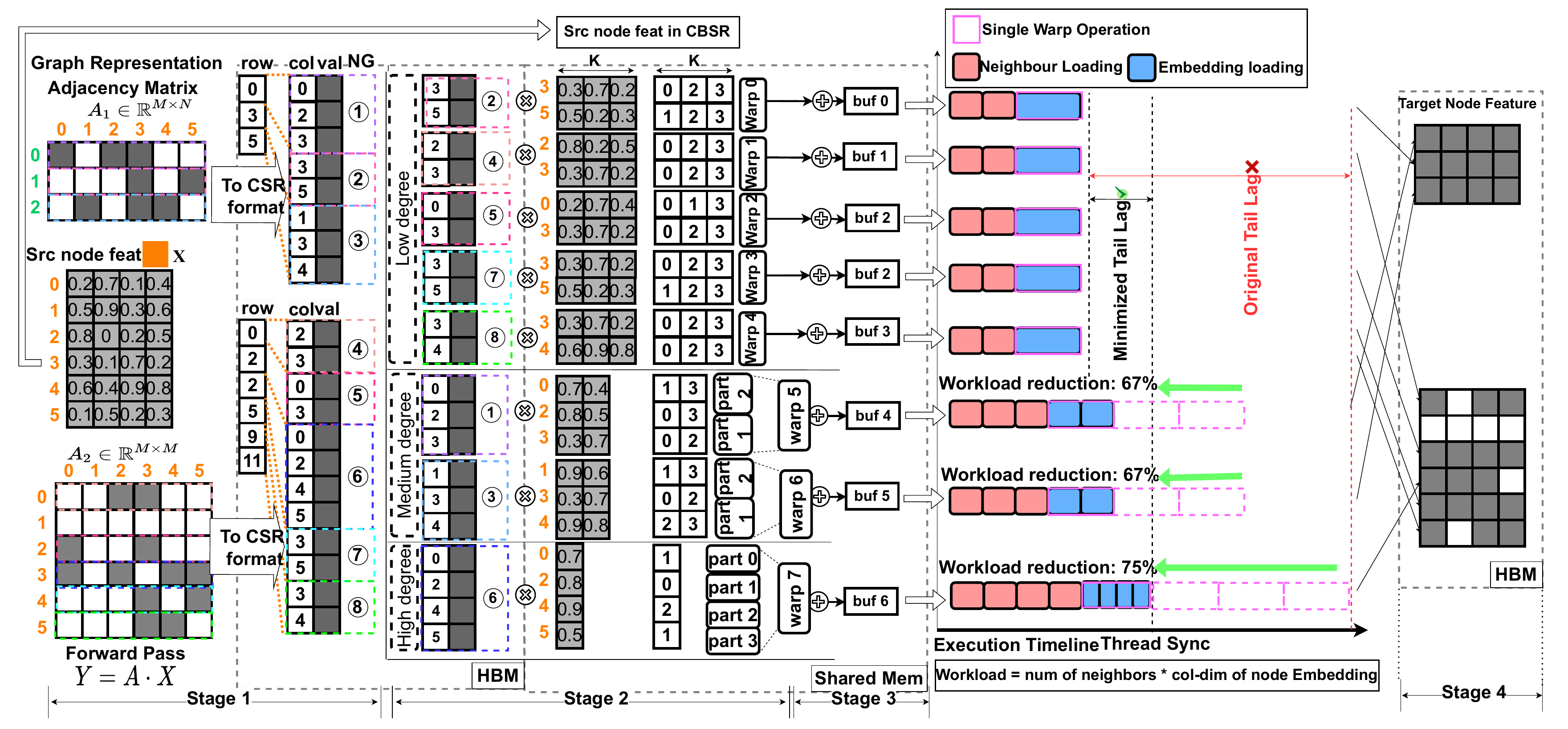}
    \caption{DR-CircuitGNN forward SpMM kernel implementation, from left to right, including adjacency matrices preprocessing at stage 1, dynamic warp partitioning at stage 2, type-specific feature aggregation at stage 3, the relative scheduling scheme, and the output processing at stage 4.}
    \label{fig:Hetero-geSpMM}
    % \vspace{-10pt}
\end{figure*}

The D-ReLU operation can be formally defined for matrix elements $A_{ij}$, where $\tau_i$ represents the row-wise threshold determined by the minimum of $k$ maximal elements selected from row $i$. This approach ensures balanced sparsity across output node embeddings while preserving the most significant features. The output comprises both the preserved element values and their relative positional indices within the embedding, enabling efficient DR-SpMM operations across different edge types during message passing.

% Dynamic threshold for heterogeneous node embeddings
\begin{equation}
\text{th}_i^{\phi_s} = \min(\text{topk}(X_{i,:}^{\phi_s}, k_{\phi_s}))
\end{equation}

% Dynamic ReLU function for heterogeneous node embeddings
\begin{equation}
f(X_{i,d}^{\phi_s}) = 
\begin{cases} 
X_{i,d}^{\phi_s}, & X_{i,d}^{\phi_s} \geq \text{th}_i^{\phi_s} \\ 
0, & X_{i,d}^{\phi_s} < \text{th}_i^{\phi_s}
\end{cases}
\end{equation}
Given there are $\psi$ types of edges, and $\phi_s$ types of nodes, $th$ is the row-wise threshold defining row-sparsity. For example, in Circuitnet, $\psi\in\{pins,near,pinned\}$ contains all available types of edges, while $\phi_s\in\{cell,net\}$ refers to all node types available.

From the output of D-ReLU, including the values of preserved elements and the relative indices of their positions in the embedding, DR-SpMM can efficiently perform balanced and rapid row-wise multiplication between the adjacency matrix per edge type and the corresponding node embeddings for the aggregation of MP.

Our redesigned DR-SpMM implementation leverages warp-level row-major and column-major encoding of adjacency matrices for forward-pass and backward-pass within kernel computations, facilitating both intra-node homogeneous SpMM and cross-type heterogeneous SpMM operations. As a simplified example shown in Figure~\ref{fig:Hetero-SpMM}, the framework first accommodates varying embedding dimensions across node types, with different K-values applied to $net$ and $cell$ nodes to generate respective CBSR (Compressed Balanced Sparse Row) encoded values and indices. Two types of node embedding in CBSR format, as well as the respective edge relations in adjacency matrices, are both leveraged during the forward pass and the backward pass.

\subsection{DR-SpMM Forward-Pass for HGNN}
Our heterogeneous forward-pass General Sparse Matrix Multiplication (GeSpMM) kernel design follows the row-wise product style as in \cite{srivastava2020matraptor} and the general format:

\begin{equation}
\begin{split}
    Y^{\phi_t} &= \sum_{\psi \in \Psi^{\phi_s \rightarrow \phi_t}} A^{\psi} \cdot X^{\phi_s} \cdot W^{\psi}, \\
    &\text{Where } A^{\psi} \in \mathbb{R}^{M_{\phi_t} \times N_{\phi_s}},\ X^{\phi_s} \in \mathbb{R}^{N_{\phi_s} \times D_{\phi_s}}, \\
    &W^{\psi} \in \mathbb{R}^{D_{\phi_s} \times D_{\phi_t}}
\end{split}
\end{equation}
Note that there are $\phi_s$ types of source nodes. $\psi$ types of edges,  $N_{\phi_s}$ denotes the number of source nodes, $M$ denotes the number of destination nodes, $Y$ is the updated node embedding, $A$ means the adjacency matrix, and $X$ is the node embedding from the last iteration. The step-by-step neighbor aggregation bore within the forward pass has the form:

\begin{align}
    Y^{\phi_t}_i &= \sum_{\psi \in \Psi^{\phi_s \rightarrow \phi_t}} \sum_{j \in \mathcal{N}^{\psi}(i)} A^{\psi}_{i,j} \cdot X^{\phi_s}_j \cdot W^{\psi} \label{eq:hetero_forward} 
\end{align}

Where $\mathcal{N}^{\psi}(i) = \{ j \mid A_{i,j}^{\psi} \neq 0 \}$ represents the neighbor group (NG) of the node $i$,  $A_{i,j}^{\psi} \in \mathbb{R}^+ \quad \text{(Edge)}$ represents the adjacency connection (edge) from source node $j$ to destination node $i$. $X_j^{\phi_s} \in \mathbb{R}^{D}$ is the corresponding embedding of source node $j$ to be aggregated onto the destination node $i$. When the equations are applied to circuit graphs from CircuitNet:

\begin{align}
    % Cell node updates (target type: cell)
    Y^{\text{cell}}_i &= \sum_{\psi \in \{\text{pinned}, \text{near}\}} 
        \sum_{j \in \mathcal{N}^{\psi}(i)} 
        A^{\psi}_{i,j} \cdot X^{\phi_s}_j \cdot W^{\psi} 
        \label{eq:cell_update} \\
    % Net node updates (target type: net)
    Y^{\text{net}}_k &= \sum_{\psi \in \{\text{pins}\}} 
        \sum_{j \in \mathcal{N}^{\psi}(k)} 
        A^{\psi}_{k,j} \cdot X^{\phi_s}_j \cdot W^{\psi} 
        \label{eq:net_update} 
\end{align}

Where $\phi_s = \phi_t \in \{cell,net\}$. The detailed algorithm with a simple example is given in Alg. 1.

Practically, for the case where node type $cell$ will receive two pieces of updated node embedding from $cell$ itself and $net$ along edge types $near$ and $pins$, respectively, the form of forward-pass message-passing then becomes:
\begin{align}
    Y^{\text{cell}} &= \max\left( 
        A^{\text{near}} \cdot X^{\text{cell}}, \ 
        A^{\text{pinned}} \cdot X^{\text{net}} 
    \right) \\
    Y^{\text{net}} &= A^{\text{pin}} \cdot X^{\text{cell}}
\end{align}

\begin{algorithm}[t]
% \footnotesize
\caption{DR-SpMM Forward-Pass}
\label{alg:hetero_forward}
\begin{algorithmic}[1]
\Function{DR-SpMM Forward-Pass}{}
\State \textcolor{blue}{\textbf{Stage 1: Adjacency Matrix Preprocessing}}
\State Encode adjacency matrices $\{A^{\psi}\}$ for all edge types $\psi \in \Psi^{\phi_s \rightarrow \phi_t}$ into CSR
\State Partition CSR into warp-level neighbor groups:
\For{each edge type $\psi$ and row $r$ in CSR}
    \State Assign warp $W_r^{\psi} \gets$ row $r$'s neighbors for relation $\psi$
\EndFor

\State \textcolor{blue}{\textbf{Stage 2: Dynamic Warp Partitioning}}
\State Classify neighbor groups by degree with warp size 32, $K_{1}$>$K_{2}$>$K_{3}$:
\begin{itemize}
    \item Low degree : No partition or into $\lceil 32/K_{1} \rceil$ parts
    \item Medium degree : Partition into $\lceil 32/K_{2} \rceil$ parts
    \item High degree : Partition into $\lceil 32/K_{3} \rceil$ parts
\end{itemize}

\State \textcolor{blue}{\textbf{Stage 3: Type-specific Feature Aggregation}}
\For{each target node type $\phi_t \in \Phi$}
    \For{each edge type $\psi \in \Psi^{\phi_s \rightarrow \phi_t}$ connecting to $\phi_t$}
        \For{each warp $W$ in partitioned warps for $\psi$}
            \State Load features $X^{\phi_s}_j$ from HBM with CBSR idx
            \State Compute: $Y^{\phi_t}_i += \sum_{j \in \mathcal{N}^{\psi}(i)} A^{\psi}_{i,j} \cdot (X^{\phi_s}_j \cdot W^{\psi})$
        \EndFor
    \EndFor
\EndFor

\State \textcolor{blue}{\textbf{Stage 4: Output Processing}}
\For{each target node type $\phi_t \in \Phi$}
    \State Write $Y^{\phi_t}$ to HBM
\EndFor
\State Preserve type-specific CBSR indices for backward pass
\EndFunction
\end{algorithmic}
\end{algorithm}

\begin{algorithm}[t]
% \footnotesize
\caption{DR-SpMM Backward-Pass}
\label{alg:hetero_backward}
\begin{algorithmic}[1]
\Function{DR-SpMM Backward-Pass}{}
\State \textcolor{blue}{\textbf{Stage 1: Gradient Preparation}}
\For{each edge type $\psi \in \Psi^{\phi_s \rightarrow \phi_t}$}
    \State Transpose $A^{\psi}$ to CSC format
    \State Reuse preserved type-specific CBSR indices from forward pass
\EndFor

\State \textcolor{blue}{\textbf{Stage 2: Type-specific Reverse Aggregation}}
\For{each source node type $\phi_s \in \Phi$}
    \For{each edge type $\psi \in \Psi^{\phi_s \rightarrow \phi_t}$}
        \For{each warp $W$ as partitioned in forward}
            \State Load gradients $\frac{\partial L}{\partial Y^{\phi_t}}$ from HBM
            \State $\frac{\partial L}{\partial X^{\phi_s}_j} += \sum_{i \in \mathcal{N}^{\psi}(j)} A^{\psi}_{i,j} \cdot \frac{\partial L}{\partial Y^{\phi_t}_i} \cdot (W^{\psi})^T$
        \EndFor
    \EndFor
\EndFor

\State \textcolor{blue}{\textbf{Stage 3: Gradient Accumulation}}
\For{each source node type $\phi_s \in \Phi$}
    \For{each source node $j$ of type $\phi_s$}
        \State Atomically add: $\frac{\partial L}{\partial X^{\phi_s}_j} \gets \sum_{\psi} \frac{\partial L}{\partial X^{\phi_s}_j}^{\psi}$
    \EndFor
\EndFor
\EndFunction
\end{algorithmic}
\end{algorithm}
In Alg. 1, the kernel implementation is further illustrated in Figure~ \ref{fig:Hetero-geSpMM}. We created an example of a heterogeneous graph where type-1 nodes pass their features along two edge relations in terms of adjacency matrices to type-2 nodes and themselves, respectively. In the first stage, the adjacency matrices will be encoded into a compressed sparse row format and further partitioned into warp-level, where each warp is responsible for one row's neighbor information, namely, the neighbor group. This step as the stage 1 refers to the line $2$ to line $7$ in Alg.1. In the next stage shown at line 9 in Alg.1, all neighbor groups are then divided into three cases in this scenario: low degree (two neighbors per node), medium degree (three neighbors per node), and large degree (four neighbors per node). Concerning the size of the neighbor group, a neighbor-size-aware scheduling technique is applied: After analysis of the NG size distribution for every edge type of every graph, D-ReLU will apply respective K-values to the NGs with respect to their sizes. Their relative warps will be partitioned accordingly. The more neighbors the NGs have, the fewer features per neighbor are required to pass, which corresponds to smaller K-values and thus fewer values per row of node features to load, and more parts per warp will be partitioned into and handle the embeddings linked to the NGs. In this manner, the need for warps does not grow linearly with the size of NGs, reducing repeated calls of the same warp from the same block of threads to process other rows' message-passing and curtail the tail lag effect caused by unadjusted, uniform warp size if applied to large-degree NGs. 
For example, in Figure \ref{fig:Hetero-geSpMM}, during the stage when each warp loads the neighbors' information and the relative embeddings, the workload of neighbor loading (red block) matches the number of neighbors for each NG, given varied sizes of NGs: 2 neighbors for small degree NG \textcircled{2},\textcircled{4},\textcircled{5},\textcircled{7},\textcircled{8}; 3 neighbors for medium degree NG \textcircled{1},\textcircled{3}; 4 neighbors for high-degree NG \textcircled{6}. Knowing the needed information of NGs' sizes, D-ReLU adjusts the K-values, indicating the preserved number of embeddings per neighbor to 
$\frac{2}{3}$ for medium-sized NGs, and $\frac{1}{3}$ for large-sized NGs of the biggest allowed K-value 3. According to the K-values adjusted, the loading of the
% pruned 
sparsified
embedding (blue block) are divided into 2 parts, even 4 parts when the NG has 3 and 4 neighbors respectively, so that their assigned warps only need to be called one time to handle the row-wise product and neighbor aggregation with fetched neighbor information, including row indices and the values indexed, as well as the corresponding rows of node embeddings.
After the neighbor aggregation (as shown at the stage 3, line 11 to line 18), the DR-SpMM forward kernel stores the result in shared memory, which is then written back to the output on HBM (High Bandwidth Memory) with information of rows' pointers of the target node, finishing the Forward Pass. Note that the indices of CBSR-encoded node features will be preserved and reused for the backward pass. This stage corresponds from line 19 to line 24 in Alg.1.

\subsection{DR-SpMM Backward-Pass}
Following the forward pass, the corresponding backward pass, supplementing the training flow is defined by kernel implementation in Alg. 2 with a general form:

\begin{equation}
\begin{split}
    \frac{\partial L}{\partial X^{\phi_s}} &= \sum_{\psi \in \Psi^{\phi_s \rightarrow \phi_t}} (A^{\psi})^\top \cdot \frac{\partial L}{\partial Y^{\phi_t}} \cdot (W^{\psi})^\top, \\
    &\text{Where } (A^{\psi})^\top \in \mathbb{R}^{N_{\phi_s} \times M_{\phi_t}}
\end{split}
\end{equation}

Where $L$ is the loss function, and $X$ and $Y$ are the node embedding from the last iteration and the current iteration, respectively. $A^\top$ implies the transposed edge connections defined by the relative adjacency matrix, with the dimension $N \times M$, which corresponds to the opposite direction of gradient passing to the forward pass.
If step by step, the backward gradient aggregation takes the form:

\begin{align}
    \frac{\partial L}{\partial X^{\phi_s}_j} &= \sum_{\phi_t} \sum_{\psi \in \Psi^{\phi_s \rightarrow \phi_t}} \sum_{i \in \mathcal{N}^{\psi}(j)} \frac{\partial L}{\partial Y^{\phi_t}_i} \cdot A^{\psi}_{i,j} \cdot (W^{\psi})^\top
\end{align}

Following the annotations used and practically, in the forward-pass,  DR-CircuitGNN already adopted element-wise $max()$ to merge the two updated node embedding of node type $cell$ along edge type $pin$ and $near$ from $cell$ and $net$ combined, the corresponding backward-pass message-passing becomes:

\begin{align}
    \frac{\partial L}{\partial X^{\text{cell}}} &= 
        \left(A^{\text{near}}\right)^\top \cdot \left( M \odot \frac{\partial L}{\partial Y^{\text{cell}}} \right) 
        + \left(A^{\text{pin}}\right)^\top \cdot \frac{\partial L}{\partial Y^{\text{net}}} \\
    \frac{\partial L}{\partial X^{\text{net}}} &= 
        \left(A^{\text{pinned}}\right)^\top \cdot \left( (1 - M) \odot \frac{\partial L}{\partial Y^{\text{cell}}} \right)
\end{align}
Where:
\begin{equation}
    M_{i,d} = \begin{cases}
        1 & \text{if } (A^{\text{near}} \cdot X^{\text{cell}})_{i,d} \geq (A^{\text{pinned}} \cdot X^{\text{net}})_{i,d} \\
        0 & \text{otherwise}
    \end{cases}
\end{equation}

\begin{figure}[t]
     \centering
     \includegraphics[width=1.0\linewidth]{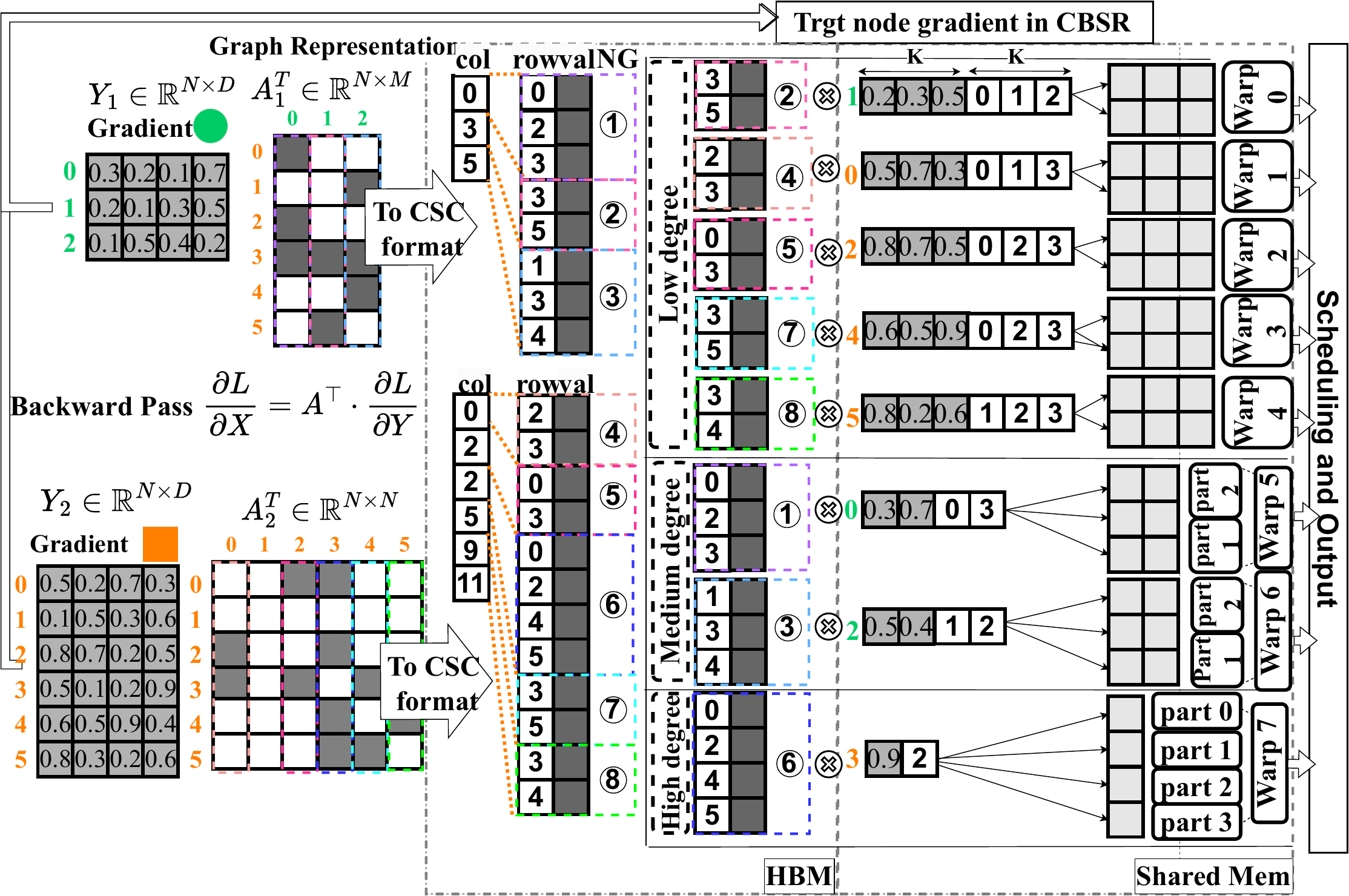}
     \caption{DR-SpMM Backward kernel.}
     \label{fig:Hetero-SSpMM}
     % \vspace{-10pt}
\end{figure}

\begin{figure}[htbp]
    \centering
    \begin{subfigure}[b]{\columnwidth}
        \includegraphics[width=\linewidth]{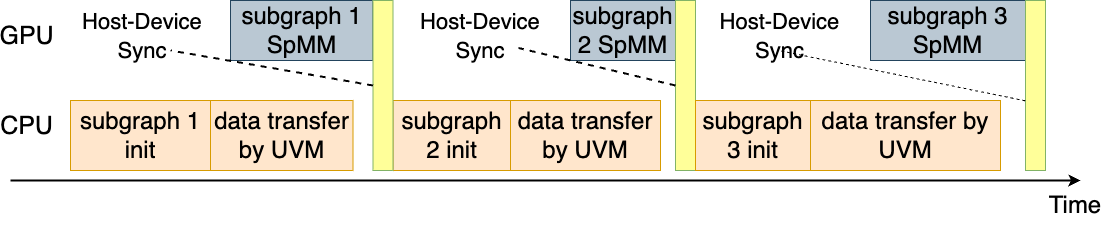}
        \caption{Timeline of HGNN}
        \label{fig:timeline1}
    \end{subfigure}
    \hfill
    \vspace{3pt}
    \begin{subfigure}[b]{\columnwidth}
        \includegraphics[width=\linewidth]{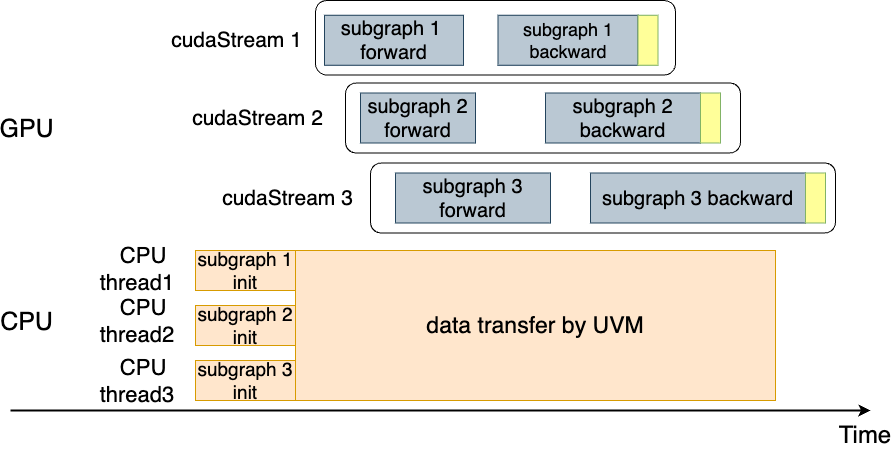}
        \caption{Timeline of our parallel design}
        \label{fig:timeline2}
    \end{subfigure}
    \caption{The final parallelism effects in DR-CircuitGNN. The initialization of subgraphs is handled in parallel by three CPU threads. The GPU kernels for different subgraphs are put into different cudastreams and then launched at the same time. The CUDA runtime automatically controls GPU resource scheduling.}
    \label{fig:timeline}
\end{figure}

at the stage 2, line 7 to line 15 in Alg.2, unchanged in content.  On the other hand, the target node gradient values are indexed by the CBSR's indices preserved in the forward pass. At the stage 2, line 8 to line 15 of Alg. 2, when the kernel is traversing NG's column-major row indices and their relative edge values, per each edge value, warps meanwhile are responsible for loading the corresponding rows of gradients from both types of target nodes to perform the column-wise product, and mapping the resulting values to the output rows using the row indices of edge values indicated in line 16 to line 21 of Alg. 2. Note that at during this stage, the NGs, node features, CBSR indices, and the results are stored on HBM while the interim product results are placed on shared memory. Similarly to the forward pass case, warps have bigger partitions for small NGs, while smaller partitions are assigned to warps handling large NGs to mitigate the tail lag throughout the process. In addition, as the source type-1 nodes receive two gradient results, they go through post-processing provided by torch, which is unrelated to the design.

\subsection{Parallel Optimization}

As illustrated in Figure~\ref{fig:Hetero-SpMM}, the circuit graph consists of two distinct node types—\textit{Net} and \textit{Cell}—and two edge types: undirected edges between \textit{Cells} and directed edges connecting \textit{Net} and \textit{Cell}. To effectively model these heterogeneous relationships during HGNN training, three separate adjacency matrices (denoted as \textcircled{1}, \textcircled{2}, and \textcircled{3} in Figure~\ref{fig:Hetero-SpMM}) are derived from the original graph. These matrices, which encode the topological relationships among different node and edge types, are preprocessed and stored as three independent subgraphs prior to training. Notably, each partition of the full circuit graph generates its own set of three subgraphs, ensuring that the heterogeneous structural information is preserved at both the global and partition levels.

The Fig.~\ref{fig:timeline1} shows the timeline of the original HGNN workflow. After the activation layer, the program sequentially loads the three subgraphs into GPU via Unified Virtual Memory(UVM)\cite{uvm}, so that the data transfer will be automatically handled by CUDA runtime and overlap with computing, and then performs forward and backward computations on each of them. All initialization tasks, including data loading, memory allocation, and host-to-device data transfer, are handled by the CPU. Crucially, both the CPU-side initialization and GPU-side kernel execution for the three subgraphs are entirely independent. Leveraging this inherent parallelism, we employ a multi-threaded CPU implementation to concurrently initialize the three subgraphs, followed by the use of three separate \textit{cudaStreams} to manage their corresponding GPU kernels. Figure~\ref{fig:timeline2} shows the timeline of our parallel design. These kernels are launched concurrently, with their execution overlapping fully when GPU resources are available or partially when resource contention occurs. Even in the worst-case scenario, where the three workloads are executed sequentially due to each fully utilizing the GPU's computing units, the use of \textit{cudaStreams} still reduces unnecessary explicit system synchronization overhead, which has been shown to significantly degrade GPU efficiency\cite{subway,liberator,emogi}. 

\section{Experiments and Analysis}

We evaluate DR-CircuitGNN through comprehensive benchmarking, focusing on runtime performance analysis of the DR-SpMM kernel (both forward and backward passes) across varying K-values and embedding dimensions of different node types of various graphs in the dataset. Our analysis also examines the impact of K-values on training efficiency and accuracy metrics to determine optimal configurations for heterogeneous graph datasets.
\subsection{Experimental Setup}
\textbf{Platform:} All experiments were conducted on an AMD EPYC 7763 64-Core Processor server with 504GB RAM and a single NVIDIA A6000-48GB GPU, using CUDA toolkit 12.6. The GPU kernels are coded in CUDA C. The DGL toolkit version is 2.2.1 with Pytorch 2.1.2, and Python 3.10.

\begin{table*}[!t]
\footnotesize
\caption{Statistics of three representative circuit designs of different sizes (i.e., small, medium, large).}
\label{tab:CircuitNet-Example}
\centering
\begin{tabular}{|c|c|c|c|c|c|c|c|c|}
\hline
 & graph id & nodes-net & nodes-cell & edges-pinned & edges-near & edges-pins & total nodes & total edges \\ \hline
\multirow{2}{*}{Design 9282-zero (Small)}  & 0 & 4628 & 7767 & 10013 & 338050 & 10013 & 12395 & 358076 \\ \cline{2-9} 
                                      & 1 & 3269 & 7347 & 7580  & 282216 & 7580  & 10616 & 297376 \\ \hline
\multirow{3}{*}{Design 2216-RISCY (Medium)} & 0 & 5331 & 9493 & 12382 & 432187 & 12382 & 14824 & 456951 \\ \cline{2-9} 
                                      & 1 & 7271 & 9733 & 18814 & 444258 & 18814 & 17004 & 481886 \\ \cline{2-9} 
                                      & 2 & 6461 & 9590 & 19227 & 409581 & 19227 & 16051 & 448035 \\ \hline
\multirow{4}{*}{Design 7598-zero (Large)}  & 0 & 5883 & 9816 & 16605 & 455383 & 16605 & 15699 & 488593 \\ \cline{2-9} 
                                      & 1 & 6183 & 9399 & 17394 & 449466 & 17394 & 15582 & 484254 \\ \cline{2-9} 
                                      & 2 & 9100 & 9579 & 34748 & 440481 & 34748 & 18679 & 509977 \\ \cline{2-9} 
                                      & 3 & 7146 & 9341 & 22056 & 483638 & 22056 & 16487 & 527750 \\ \hline
\end{tabular}
\end{table*}

\textbf{Datasets:} 

In order to evaluate DR-CircuitGNN on the large-scale EDA graph dataset, we first implement SpMM performance assessment with size-stratified CircuitNet samples (Table \ref{tab:CircuitNet-Example}); and comprehensive testing on CircuitNet \cite{yang2022versatile,10158384,2024circuitnet}, an open-source, extensive EDA dataset encompassing over 10,000 circuit designs. For comparative analysis with canonical GNN architectures, we constructed Mini-CircuitNet, comprising 120 randomly sampled designs (100 training, 20 testing) from CircuitNet, processed according to the protocol from \cite{10158384}.
More importantly, to further investigate heterogeneous K-values with regard to the two node types' embeddings plus the relative edges and the performance of DR-CircuitGNN, we use all of the designs on full-size CircuitNet and profile the performance in correlation scores. This dataset features enhanced representation of circuit designs dedicated to EDA and computer-aided design (CAD) based on very large-scale integration (VLSI) \cite{yang2022versatile,10158384,2024circuitnet}. With more than 10,000 samples extracted from commercial design tools and six open-source RISC-V designs, these designs can be applied to typical cross-stage prediction tasks, such as routability prediction and IR drop prediction, where in this work we focus on the routability or congestion prediction that again highlights rank correlation among graphs over absolute error in value prediction.  Owing to the memory constraint to generalize the network-wise test to other canonical homogeneous GNNs, we randomly sample a small subset of 120 designs from the full CircuitNet to make a Mini-CircuitNet, where 100 of them are used for training and the rest 20 designs are used for testing. The dataset pre-preprocessing follows \cite{10158384} to fit in both formats of homogeneous graphs and heterogeneous graphs. 

\textbf{Models and Configurations:} 

The SpMM kernel benchmarking is compared against the DGL implementation cuSPARSE \cite{naumov2010cusparse}, and the SpMM kernel from  GNNAdvisor (GNNA) \cite{wang2021gnnadvisor}, which was optimized specifically for GNN. We profiled the optimal K-value for each subgraph in preprocessing, which will take about 20 minutes for the whole dataset, while the end-to-end training time is dozens of hours.  According to our performance breakdown in Section \ref{sec:breakdown}, our DR-SpMM can bring up to 39\% time saving, which is about 1 - 4 hours in end-to-end training, far exceeding the cost of profiling. Besides, the profiling is a one-time effort for one dataset. The detail about how to profile the optimal $K$ is illustrated in Section \ref{sec:4.3}.

For CircuitNet evaluation, we implemented both heterogeneous (two layers of HeteroGraphConv with Graph Convolution Network (GCN) and dual GraphSAGE submodules) and homogeneous (three-layer GraphSAGE, GCN, and Graph Attention Network (GAT) \cite{velickovic2017graph}) approaches. With the homogenous GNN model for the CircuitNet dataset, GraphSAGE, GCN, and GAT \cite{velickovic2017graph} will be applied to compare with DR-CircuitGNN's performance. Three metrics are mainly referred to, Pearson, Kendall, and Spearman scores, which weigh more in rank correlation recognition and are more favored in EDA design applications \cite{yang2022versatile}. 
We followed the optimal hyperparameter setup\cite{yang2022versatile} for GraphSAGE, GCN, and GAT on the CircuitNet dataset.
Specifically, all the baseline models are trained with three layers for 50 epochs using a learning rate of 0.001 and weight decay of 0.0002. The GraphSAGE will be adjusted to the 'mean' mode. Our DR-CircuitGNN was configured with two layers, using a learning rate 0.0002 and weight decay of 0.00001, which was the optimal setup on the CircuitNet dataset in our experiments. Notably, it contains approximately twice as many parameters as GAT, SAGE, and GCN, with training time scaling proportionally.

\begin{table}[!t]
\caption{Congestion prediction results comparison between Homogeneous and our DR-CircuitGNN on Mini-CircuitNet.}
\label{tab:DR-CircuitGNN-and-GNNs-on-CircuitNet}
\centering
\resizebox{0.98\linewidth}{!}{
\begin{tabular}{|l|ccccc|}
\hline
Model & Pearson & Spear. & Ken. & MAE & RMSE \\ 
\hline
GCN & 0.347 & 0.493 & 0.372 & \textbf{0.027} & \textbf{0.033} \\
SAGE & 0.347 & 0.494 & 0.373 & 0.027 & \textbf{0.033} \\
GAT & 0.347 & 0.494 & 0.373 & \textbf{0.027} & \textbf{0.033} \\
\textbf{DR-CircuitGNN (ours)} & \textbf{0.442} & \textbf{0.511} & \textbf{0.384} & 0.043 & 0.098 \\
\hline
\end{tabular}}
\vspace{-5pt}
\end{table}

\begin{table}[t]
\centering
\resizebox{\columnwidth}{!}{%
\begin{tabular}{l|c|c|cc|cc}
\hline
\multirow{2}{*}{\textbf{Design Name}} & 
\multirow{2}{*}{\textbf{Graph ID}} &
\multirow{2}{*}{\textbf{Dim}} &
\multicolumn{2}{c|}{\textbf{Speedup vs. cuSPARSE}} &
\multicolumn{2}{c}{\textbf{Speedup vs. GNNA}} \\
\cline{4-7}
 & & & \textbf{Forward} & \textbf{Backward} & \textbf{Forward} & \textbf{Backward} \\
\hline
\multirow{6}{*}{2216-RISCY} 
 & \multirow{2}{*}{graph0} & 64 & 2.61 & 2.63 & 10.43 & 10.50 \\
 & & 128 & 2.35 & 2.45 & 9.89 & 10.29 \\
\cline{2-7}
 & \multirow{2}{*}{graph1} & 64 & 2.84 & 2.88 & 12.08 & 12.24 \\
 & & 128 & 2.54 & 2.53 & 10.80 & 10.75 \\
\cline{2-7}
 & \multirow{2}{*}{graph2} & 64 & 2.96 & 3.09 & 12.12 & 12.64 \\
 & & 128 & 2.39 & 2.41 & 10.19 & 10.27 \\
\hline
\multirow{8}{*}{7598-zero-riscy} 
 & \multirow{2}{*}{graph0} & 64 & 2.64 & 2.56 & 10.74 & 10.42 \\
 & & 128 & 2.28 & 2.29 & 9.75 & 9.79 \\
\cline{2-7}
 & \multirow{2}{*}{graph1} & 64 & 2.63 & 2.62 & 10.74 & 10.67 \\
 & & 128 & 2.45 & 2.46 & 10.37 & 10.41 \\
\cline{2-7}
 & \multirow{2}{*}{graph2} & 64 & 2.89 & 2.89 & 11.77 & 11.77 \\
 & & 128 & 2.73 & 2.82 & 11.52 & 11.93 \\
\cline{2-7}
 & \multirow{2}{*}{graph3} & 64 & 2.62 & 2.51 & 10.64 & 10.20 \\
 & & 128 & 2.48 & 2.52 & 10.36 & 10.54 \\
\hline
\multirow{4}{*}{9282-zero-riscy} 
 & \multirow{2}{*}{graph0} & 64 & 2.59 & 2.57 & 10.76 & 10.67 \\
 & & 128 & 2.33 & 2.43 & 9.95 & 10.38 \\
\cline{2-7}
 & \multirow{2}{*}{graph1} & 64 & 2.57 & 2.62 & 10.66 & 10.85 \\
 & & 128 & 2.40 & 2.48 & 10.05 & 10.36 \\
\hline
\multirow{2}{*}{\textbf{Average Performance}} & \multirow{2}{*}{-} & 64 & \multicolumn{2}{c|}{\textbf{2.71}} & \multicolumn{2}{c}{\textbf{11.10}} \\
 & & 128 & \multicolumn{2}{c|}{\textbf{2.44}} & \multicolumn{2}{c}{\textbf{10.42}} \\
\hline
\end{tabular}
}
\caption{End-to-End performance comparison across 2 different embedding dimensions (Dim=64 and Dim=128) on 3 representative circuit designs.}
\label{tab:speedup_combined}
\vspace{-20pt}
\end{table}

\begin{figure*}
    \includegraphics[width=1.0\linewidth]{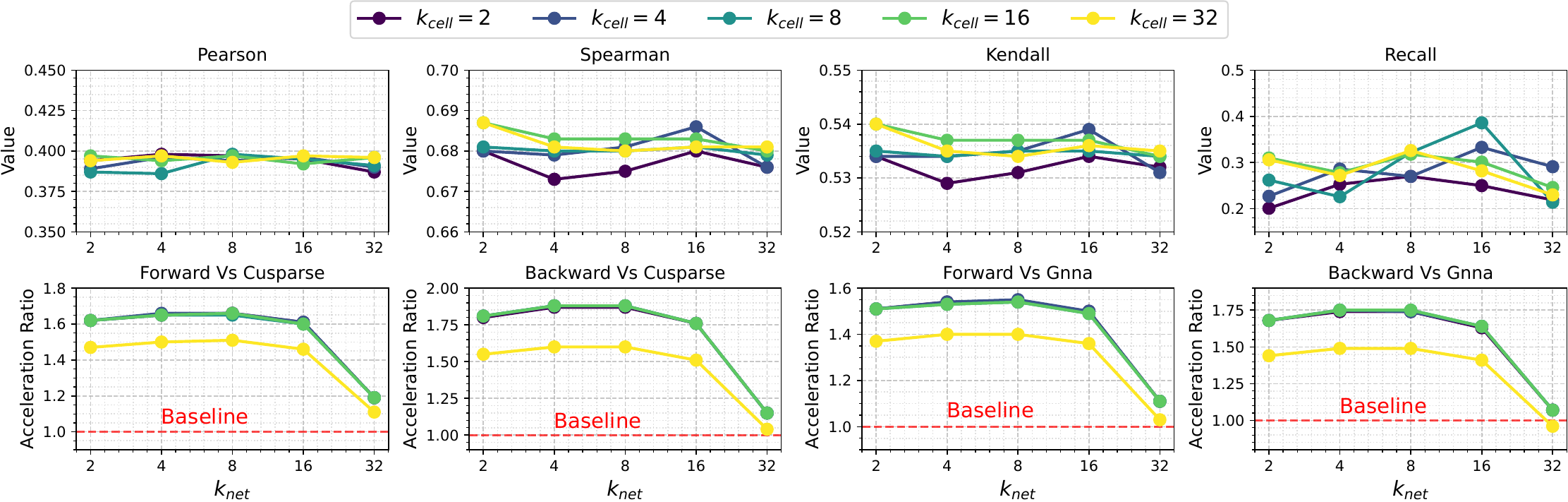}
    \caption{
    Experimental results of training with varying $K_{net}$ and $K_{cell}$ on Mini-CircuitNet. The first row is the correlation scores results and the second row shows the training speedup over baselines.
    }
    \label{fig:Kernel-100-Designs}
    \vspace{-10pt}
\end{figure*}

\subsection{DR-SpMM Forward and Backward Kernels on Heterogeneous Circuit Graphs}
\begin{figure*}[t]
    \centering
 \begin{subfigure}{\linewidth}
 \centering
          \includegraphics[width=1.0\linewidth]{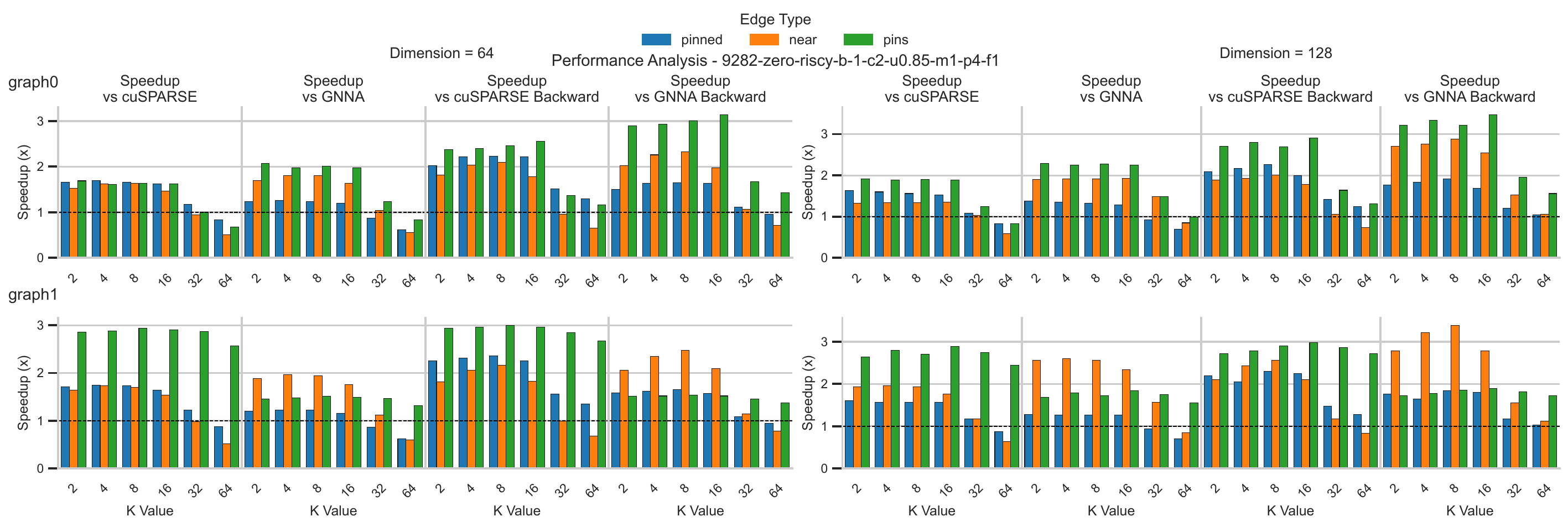}
 \end{subfigure}

  \begin{subfigure}{\linewidth}
 \centering
     \includegraphics[width=1.0\linewidth]{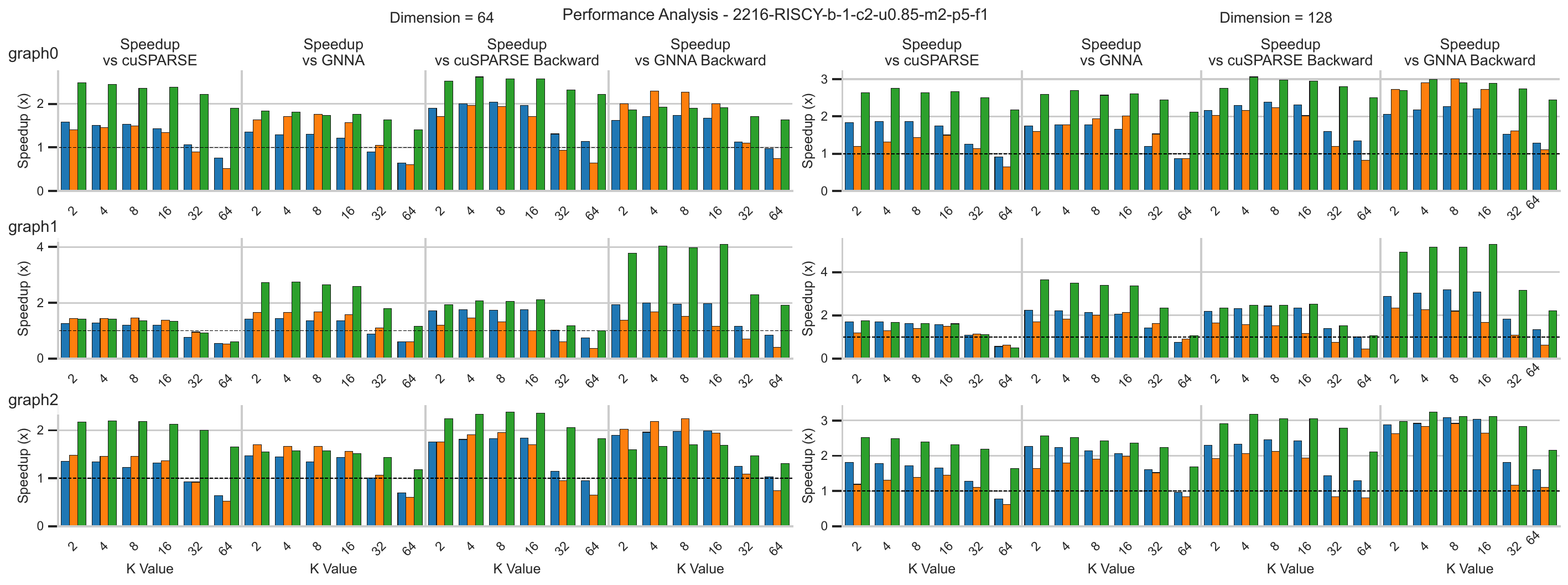}
 \end{subfigure}
 
  \begin{subfigure}{\linewidth}
 \centering
     \includegraphics[width=1.0\linewidth]{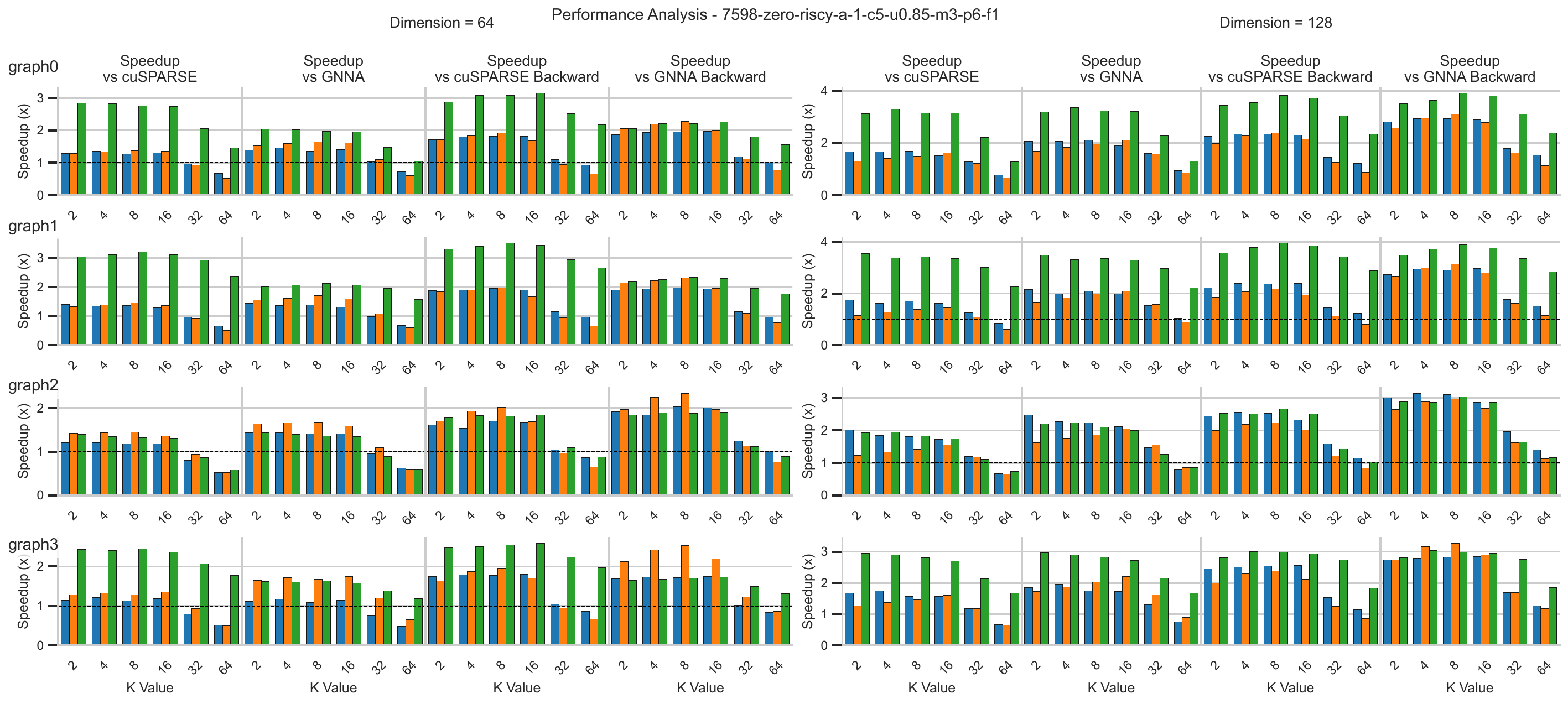}
 \end{subfigure}
 
    \caption{DR-SpMM kernels runtime speed-up under varying K against cuSPARSE and GNNA across 3 representative example circuit designs (i.e., 9282-zero, 2216-RISCY and 7598-zero) with node embedding = 64 and 128.}
    \label{fig:kernel-runtime}
    \vspace{-10pt}
\end{figure*}

The profiled results of our DR-SpMM forward and backward kernels in comparison with cuSPARSE and GNNA are shown in Figure~\ref{fig:kernel-runtime}.
Analysis of kernel performance, comparing DR-SpMM against cuSPARSE and GNNAdvisor (Figure~\ref{fig:kernel-runtime}), reveals consistent acceleration patterns when dimensionality, namely, the K value, remains below the warp thread limit of 32. The most significant gains manifest in module-wise acceleration. When the original dimension of embedding is 64, our approach has sped up in forward and backward pass by $2.75\times$ and $4.09\times$ respectively over GNNA, $3.21\times$ and $3.51\times$ over cuSPARSE. When the original dimension of the embedding is 128, our approach achieves $3.64\times$ and $5.28\times$ improvements in forward and backward passes, respectively, compared to GNNAdvisor, and $3.55\times$ and $3.98\times$ improvements versus cuSPARSE.
Performance characteristics vary notably across edge types. For example, when the dimension of embedding is 64, the best edge-wise case is on $pins$, where our DR-SpMM kernels obtain the highest speed up of $3.21\times$ in forward, and $3.51\times$ in backward over cuSPARSE, and $2.75\times$ in forward and $4.09\times$ in backward over GNNA. when the edge type is $near$, which most closely resembles canonical SpMM with square adjacency matrices, our DR-SpMM kernels has the highest speed up of $1.73\times$  and $2.16\times$ over cuSPARSE, $1.97\times$  and $2.53\times$ over GNNA, while such speed up ratio generally stays the same for cuSPARSE on edge type $pinned$ whose adjacency matrix has more columns than rows, the speed up over GNNA goes down to $1.47\times$ in forward, $2.03\times$ in backward. 
These results show our methods' consistently enhanced performance against established baselines. In addition, under comparable numbers of vertices that shape the dimensions of all three types of edges, the speed up performance present obvious variance, meaning our kernels are more in favor of adjacency matrices that has more rows than columns instead of the opposite. This shows our kernel design can benefit more from heterogeneous graphs whose source nodes are more than the target nodes.

\subsection{DR-CircuitGNN Evaluation on CircuitNet}
\label{sec:4.3}
Evaluation of DR-CircuitGNN on congestion prediction tasks shows its effectiveness over homogeneous GNN baselines. On Mini-CircuitNet, our model achieves enhanced correlation metrics with Pearson, Spearman, and Kendall scores of 0.442, 0.511, and 0.384, respectively. The full-scale CircuitNet implementation further validates these results, showing improved Spearman and Kendall scores of 0.68 and 0.535, with a marginal decrease in Pearson to 0.387. 
% \textcolor{blue}{
Performance analysis in varying $K$ values, as illustrated in Figure~\ref{fig:Kernel-100-Designs}, has revealed optimal acceleration in the range between $k_{net}=2$ and $k_{net}=8$.
% } 
The model maintains stable performance throughout the K-value range, where the acceleration reaches up to 1.65×/1.88× and 1.54×/1.75× versus DGL implementation and GNNAdvisor, which can be translated into a reduction in training time by $38.27\%/46.81\%$ and $35.1\%/42.86\%$ in forward/backward. Although the acceleration ratio decreases as values approach 32, these results demonstrate consistent performance advantages while maintaining metric stability across different configurations. Note that the rise in RMSE and MSE indicates more absolute values shifted from the original ones, yet acceptable given the great increase in rank correlation scores. 
% \textcolor{blue}{
To further maximize the end-to-end acceleration of training, we exhaustively search graph-specific optimal K-values: $k_{pinned}$, $k_{near}$, and $k_{pins}$, and run parallel message-passing SpMM under the given graph of the selected three designs.
Due to the influence of graph topology, the optimal $K$ value varies for each subgraph from different designs. Therefore, during the preprocessing phase, we profile the DR-SpMM kernel performance with different $K$ values on each subgraph. In the CircuitNet dataset, since the dimension of node embeddings for each subgraph is either 64 or 128, and we need to ensure that the number of non-zero elements remaining in node embeddings after activation functions is a power of two to maximize GPU parallel resource utilization, the candidate $K$ values are selected from powers of 2 smaller than 64, specifically 2, 4, 8, 16, 32, and 64. We measure the performance of the DR-SpMM kernel under each $K$ value, select the optimal $K$ value that delivers the best performance for each subgraph, and apply it to the end-to-end training.
Table. \ref{tab:speedup_combined} shows that with parallelism enabling message passing on all edges in parallel, our kernels have at best $2.71\times$ of end-to-end acceleration over cuSPARSE, and $11.1\times$ speedup over GNNA. When extended to the 100 training designs, the overall runtime acceleration can be at best $3.11\times$ and $11.58\times$ over cuSPARSE and GNNA, respectively.
% }

\subsection{Breakdown of the Optimizations}
\label{sec:breakdown}
\begin{figure}[t]
    \centering
    \includegraphics[width=\linewidth]{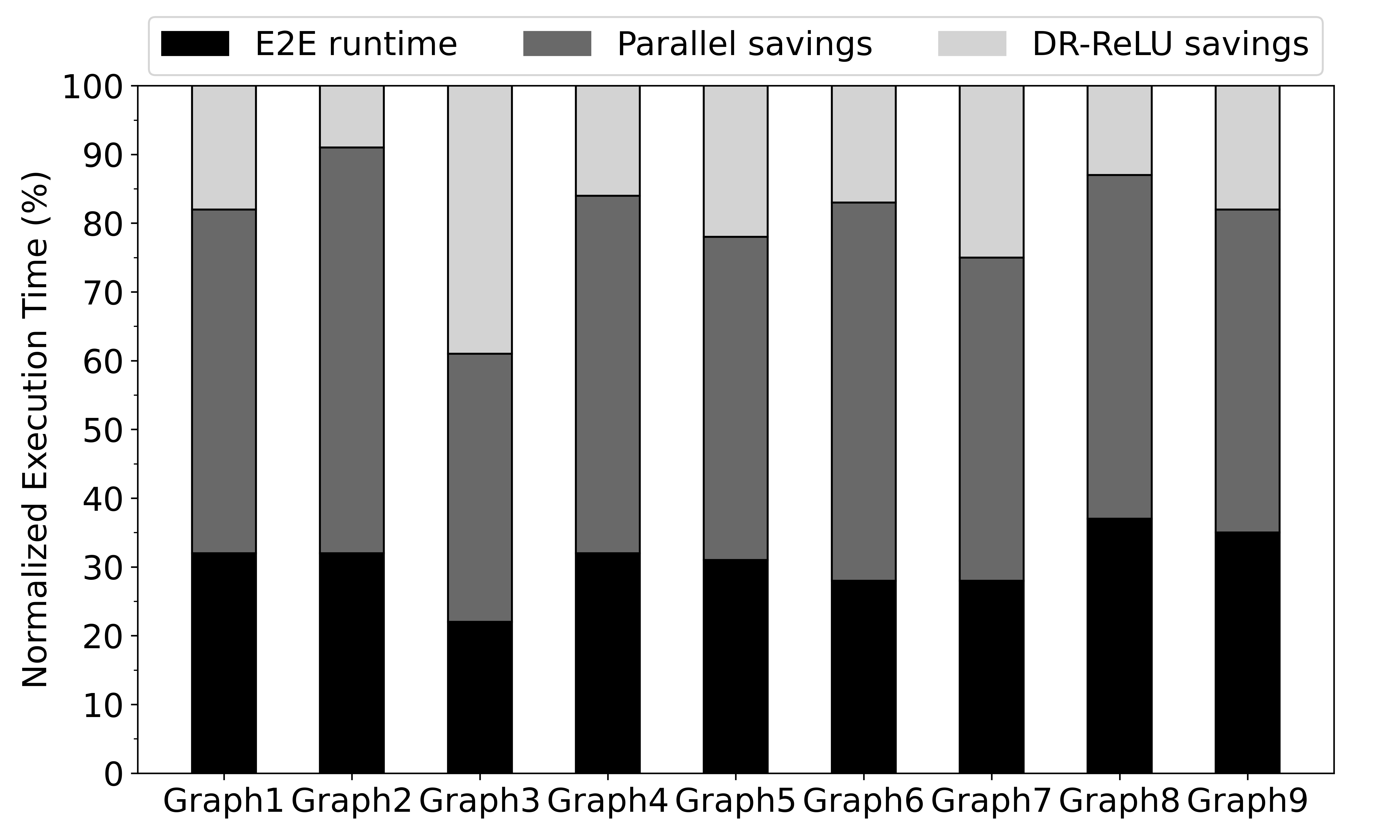}
    \caption{Breakdown of the optimization benefits on randomly selected 9 graphs in the CircuitNet dataset.}
    \label{fig:breakdown}
    \vspace{-5pt}
\end{figure}

To further understand the details of the performance improvement in End-to-End workflow, we use \textit{DR-ReLU savings} to represent the optimization brought by our optimized kernel, and \textit{Parallel savings} to represent the benefits of our parallel processing of three types of subgraphs. We use the results of cuSPARSE as the baseline. We explicitly disable parallel processing of subgraphs to measure the sole performance impact of our DR-ReLU SpMM kernel and activate the parallel scheme to measure how much improvement can be achieved from it. Figure~\ref{fig:breakdown} shows the breakdown of the performance gain on randomly selected 9 graphs from the CircuitNet dataset.

The experimental results show that our kernel optimization leads to an average of 19.3\% execution time reduction. However, the actual performance gain varies across different graphs due to two key factors: (1) the optimal 
% pruning 
sparsification threshold is graph-dependent, and (2) the inherent topological characteristics of each graph significantly influence kernel efficiency. In the best-case scenario (Graph 3), our optimization achieves a 39\% execution time reduction, indicating the effectiveness of our profiling-based threshold selection strategy. Conversely, the performance gain drops to 9\% for Graph 2, the worst-case scenario. Further profiling and topological analysis reveal that Graph2 exhibits a relatively uniform degree distribution, which limits the opportunities for our DR-ReLU optimization to exploit workload balance scheduling in the kernel computing.

Our parallel scheme achieves an average end-to-end execution time reduction of 49.6\%, with consistent performance gains across different graphs. Notably, the end-to-end evaluation encompasses not only kernel execution time but also system-level overheads, including data loading, initialization, memory allocation, and possible resource scheduling. To minimize these overheads, our design parallelizes data loading and initialization on the CPU side using three dedicated threads, each handling a separate subgraph. On the GPU side, we leverage three independent \textit{cudaStream} to launch kernels concurrently. Since these \textit{cudaStream} have no data dependencies, they can be issued simultaneously and executed in parallel by GPU computing units. However, perfect parallelism is rarely achieved due to resource contention: when one or two kernels fully utilize GPU resources, the remaining kernels execute concurrently but with only partial overlap. Additionally, CUDA runtime scheduling and context switching introduce further overhead, preventing ideal parallel execution among the three \textit{cudaStream}. If a fine-grained memory consistency scheme and fine-grained scheduling API were available to GPUs, e.g., CXL\cite{cxl}, a better \textit{cudaStream} scheduling is possible, we leave it for our future work.
\section{Related Work}
% \textcolor{red}{shuold we introduce GNNA? not sure, let @yuebo decide}
The application of GNN to circuit-related tasks has been paid attention to for a long time. Previous efforts addressed distributed circuit design\cite{pmlr-v97-zhang19e} of resonators in order to replace a traditional electromagnetic (EM) simulator. However, homogenous GNNs present less satisfying capability of interpreting EDA-related design netlists to obtain better expressiveness and thus the learning performance. Owing to the status quo, \cite{10.1145/3489517.3530675}
has applied heterogeneous graph neural networks (HGNNs) with lattice-based graph interpretation of netlist data to EDA problems. \cite{2024circuitnet} introduces CircuitNet as an open-source dataset for machine learning in VLSI CAD, offering improved domain-specific evaluation metrics and learning strategies, thereby providing a common platform for evaluating HGNN approaches in circuit analysis. Zhou\cite{relate1} et al. proposed an HGNN-based imitation learning approach for gate sizing acceleration that leverages heterogeneous circuit representations to predict candidate gates for resizing, demonstrating significant runtime reductions in iterative Lagrangian relaxation methods.

\section{Conclusion}
In this paper, we propose DR-CircuitGNN, a specialized framework for heterogeneous circuit graphs that integrates a Dynamic-ReLU mechanism, which efficiently sparsifies multiple types of node embeddings within heterogeneous graphs to inject balanced regular sparsity, customized SpMM kernels accelerating multi-edge message-passing for forward/backward passes, and parallel subgraph scheduling to accelerate end-to-end HGNN workflow. DR-CircuitGNN achieves up to 3.51× and 4.09× speedups in forward and backward propagation, respectively, while delivering up to 2.7× end-to-end training improvements over state-of-the-art methods, all with minimal accuracy loss on large-scale EDA benchmarks. These results highlight the effectiveness of dynamic
% activation pruning 
embedding sparsification
and CPU-GPU concurrency in addressing the computational demands of next-generation circuit design automation. 

\begin{acks}
    This research was supported in part by NSF SHF-2505770, Semiconductor Research Corporation (SRC) Artificial Intelligence Hardware program.
\end{acks}

\bibliographystyle{unsrt}
\bibliography{ref}

\end{document}